\documentclass[sigconf]{acmart}
\usepackage{graphicx}
\usepackage{natbib}
\usepackage{caption}
\usepackage{newfloat}
\usepackage{listings}
\usepackage{xcolor}

\usepackage{multirow}
\usepackage{multicol}
\usepackage{booktabs}
\usepackage{arydshln}
\usepackage{color}
\usepackage{makecell}
\usepackage{subcaption}
\usepackage{tabularx}
\usepackage{colortbl}
\usepackage{pifont}

\definecolor{lightgreen}{RGB}{150, 250, 170}
\definecolor{lightred}{RGB}{255, 100, 100}

\usepackage[most]{tcolorbox}
\usepackage{float}
\usepackage{xspace}
\usepackage{enumitem}
\tcbset{
  promptbox/.style={
    width=0.95\textwidth,
    colback=blue!5!gray!5,
    colframe=blue!30!gray,
    colbacktitle=blue!5!gray!5!,
    fonttitle=\bfseries\color{black},
    coltitle=black,
    boxrule=0.3mm,
    arc=3mm,
    top=6pt,
    bottom=6pt,
    left=10pt,
    right=10pt,
    enhanced,
    center,
    attach boxed title to top left={xshift=5mm, yshift=-3mm},
    boxed title style={
      arc=2mm,
      colframe=blue!30!gray,
      colback=blue!15!gray!15!
    },
    title filled,
    drop shadow={xshift=0.5mm, yshift=-0.5mm, fill=gray!30}
  }
}

\newtcolorbox{PromptBox}[2][]{promptbox, title=#2, #1}

\newcommand{\ours}{\textsc{ForestBench}}

\newcommand{\vpara}[1]{\vspace{0.05in}\textbf{#1 }}
\newcommand{\para}[1]{\vspace{0.05in}\noindent\textbf{#1 }}
\newcommand{\secref}[1]{Section~\ref{#1}}

\newcommand{\tableref}[1]{Table~\ref{#1}}

\DeclareCaptionStyle{ruled}{labelfont=normalfont,labelsep=colon,strut=off}
\floatstyle{ruled}
\newfloat{listing}{tb}{lst}{}
\floatname{listing}{Listing}
\author{Guo Chen}
\email{cg1281838223@email.swu.edu.cn}
\affiliation{%
  \institution{Southwest University}
  \city{Chongqing}
  \country{China}
}
\affiliation{%
  \institution{Tencent}
  \city{Beijing}
  \country{China}
}
\author{Ziwen Li}
\email{pique0202@email.swu.edu.cn}
\affiliation{%
  \institution{Southwest University}
  \city{Chongqing}
  \country{China}
}
\author{Reed Li}
\email{reedsli@tencent.com}
\affiliation{%
  \institution{Tencent}
  \city{Beijing}
  \country{China}
}
\author{Yu Lu}
\email{herberttli@tencent.com}
\affiliation{%
  \institution{Tencent}
  \city{Beijing}
  \country{China}
}
\author{Haibo Shi}
\email{bobsimons@tencent.com}
\affiliation{%
  \institution{Tencent}
  \city{Beijing}
  \country{China}
}
\author{Bingbing Xu}
\email{xubingbing@ict.ac.cn}
\affiliation{%
  \institution{Institute of Computing Technology, Chinese Academy of Sciences}
  \city{Beijing}
  \country{China}
}
\author{Junjie Huang}
\email{junjiehuang@swu.edu.cn}
\affiliation{%
  \institution{Southwest University}
  \city{Chongqing}
  \country{China}
}
\authornote{Corresponding author}

\title{\ours: A Unified Graph Framework for Evaluating Multi-Agent Collaboration}

\begin{document}

\begin{abstract}
Multi-agent systems (MAS) built on Large Language Models (LLMs) are proliferating rapidly, but their heterogeneous execution traces provide no common basis for evaluation across methods. Outcome-only benchmarks discard collaborations, whereas LLM-as-Judge evaluation requires additional, model-dependent inference and can vary with the LLM and rubric. We introduce a generalizable evaluation framework that maps native MAS traces into a shared space of unified collaboration graphs, enabling different methods to be evaluated under the same representation, reference set, and metric panel. Candidate graphs are compared with a query-specific reference forest. Each forest is a benchmark-provided collection of verified-success graphs: it records diverse ways in which representative MAS methods can complete the task, rather than prescribing a unique optimal process. Instantiating the framework as \ours\footnote{Data is available at \url{https://github.com/WinstonCHEN1/ForestBench}}, we filter $844$ collaboration-necessary queries from seven public datasets, precompute ten successful target-conditioned reference graphs per query, and evaluate six representative MAS frameworks. Controlled backbone, reference-construction, and perturbation studies test the stability and scope of evaluation. Once the benchmark forests are built, \ours~scores a trace in milliseconds without further LLM inference, providing a reusable structural basis for comparing diverse MAS collaboration traces.
\end{abstract}

\maketitle

\section{Introduction}

Multi-agent systems (MAS) built on Large Language Models (LLMs) have become one of the most active design paradigms for complex reasoning, software engineering, and scientific workflows~\citep{guo2024large,li2024survey,han2024llm}. Their promise comes fundamentally from collaboration: by dividing labor among planners, executors, critics, and tool users, exchanging intermediate evidence, and iteratively checking one another, MAS can organize capabilities that a single agent cannot deploy as effectively in one pass~\citep{du2023improving,wu2024autogen,hong2024metagpt}. Driven by this prospect, novel MAS frameworks are being proposed at an unprecedented pace, encompassing fixed pipelines, dynamic role assignment, and self-organizing groups of agents~\citep{chen2024agentverse,zhou2025reso}.

\begin{figure}[t]
\centering
\includegraphics[width=1\linewidth]{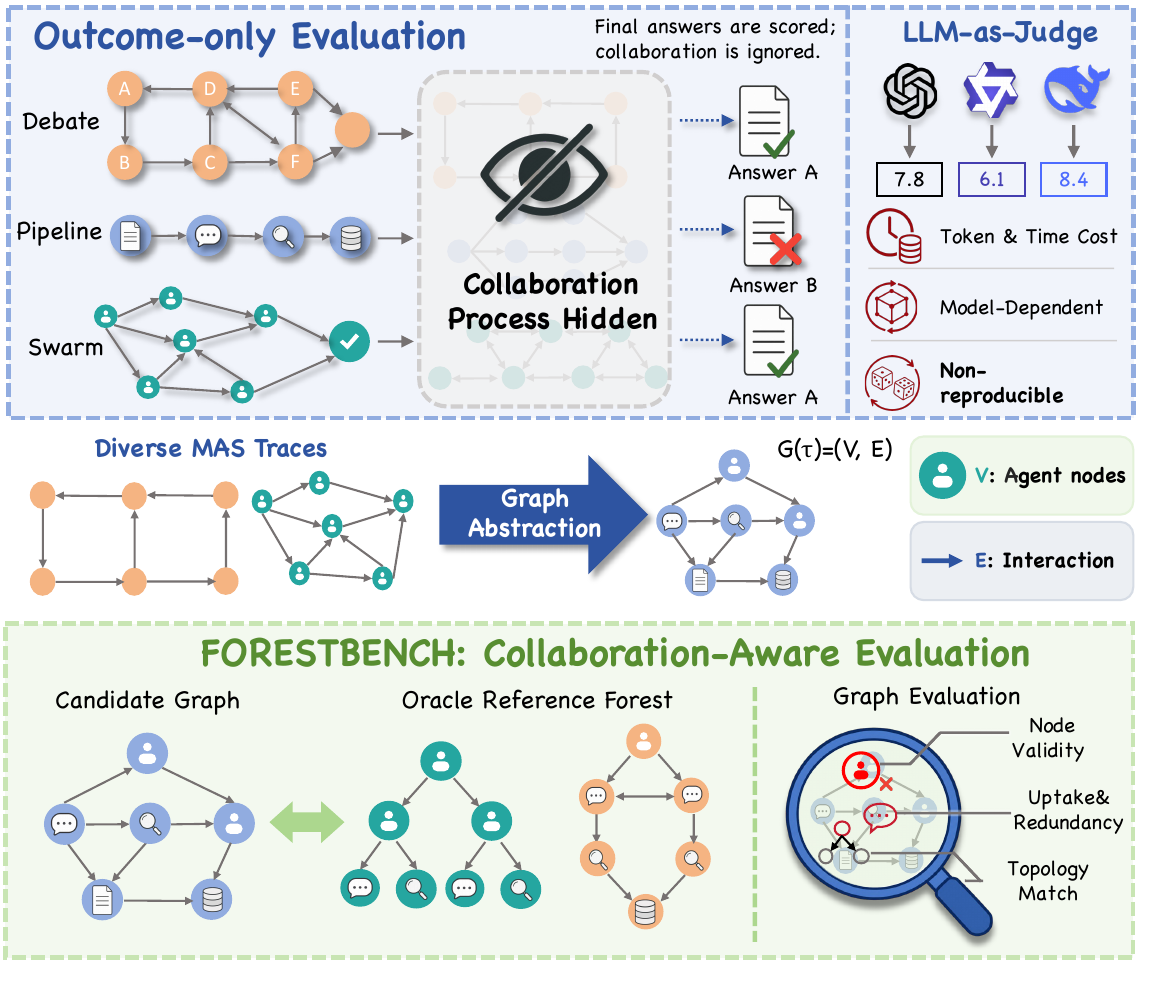}
\caption{Native traces from different MAS methods are heterogeneous, outcome-only benchmarks discard them, and LLM-as-Judge requires recurring model-dependent inference. \ours{} maps each trace into a collaboration graph, treats verified-success structures as reference trees that jointly form a reference forest, and compares candidate graphs with this forest using structural metrics.}
\label{fig:fig1}
\vspace{-10pt}
\end{figure}

However, the speed at which new MAS frameworks appear is now far outpacing our ability to evaluate and compare how they execute and collaborate. Existing studies still rely primarily on outcome benchmarks such as GSM8K, MMLU, and HumanEval, which reduce an entire MAS execution to one final answer and therefore discard the collaboration that distinguishes MAS in the first place~\citep{cobbe2021training,hendrycks2020measuring,chen2021evaluating}. Process-aware alternatives remain limited: native traces are tied to framework-specific event schemas, role vocabularies, and communication protocols, while LLM-as-Judge can read these traces only through additional model- and rubric-dependent inference with recurring cost and limited reproducibility~\citep{chan2024chateval,yehudai2025survey,chen2026contextual,cemri2026multi}. As shown in Figure~\ref{fig:fig1}, the field therefore lacks a common process-level basis for horizontal comparison: outcome-only evaluation hides collaboration, whereas free-form judging does not place heterogeneous methods in a fixed, reproducible evaluation space. What is needed is a generalizable representation that normalizes different MAS executions and supports deterministic comparison under shared criteria.

Graphs provide a natural basis for this representation~\citep{zhuge2024gptswarm,qian2025scaling}. We model a MAS execution as a collaboration graph whose nodes are agent actions and whose edges encode communicative or information-flow dependencies~\citep{zhang2025aflow,chen2024magdi,lee2025gemmas}. This abstraction removes framework-specific logging conventions while retaining who contributes, who consumes whose output, and how information moves through the system, thereby mapping otherwise incompatible traces into the same structural space~\citep{besta2025demystifying}. The graph-based view also separates representation from evaluation: task accuracy continues to measure whether the final outcome is correct, while deterministic graph statistics characterize how that outcome was produced. Yet one graph represents only one collaboration process. The same query may be solved through debate, a planner--specialist pipeline, or a peer swarm, and choosing any one as the gold process would privilege its coordination paradigm~\citep{wang2022self, yao2023tree}. This motivates the forest metaphor behind \ours{}: each verified-success collaboration structure serves as a \emph{reference tree}, recording one way in which roles and information flows grow toward a correct outcome, while structurally different reference trees form a \emph{reference forest}. A candidate method is then compared against this forest rather than one prescribed tree, preserving the plurality of successful collaboration while maintaining a fixed reference for reproducible evaluation.

In this paper, we develop this view into a concrete evaluation pipeline. We first introduce a framework that projects every MAS run to a uniform directed acyclic collaboration graph and defines a metric panel over graph structure, information uptake, redundancy, and cost. The headline metric, \emph{Forest Match}, measures the structural alignment of a candidate graph with a query-specific reference forest of verified-success behaviors. We then instantiate the framework as \ours, a benchmark built by (i) filtering $844$ collaboration-necessary queries from seven public datasets through a depth-width-decomposability pipeline, (ii) precomputing ten successful reference graphs per query from six representative MAS paradigms plus a single-agent structural anchor, and (iii) exposing adapters that convert native MAS logs into the common graph representation. The reference forests are a central released component of the benchmark: users can evaluate new MAS methods without rerunning an LLM judge or reconstructing successful traces.

Using this framework and benchmark, we evaluate six representative MAS methods with diverse backbone LLMs and conduct controlled backbone and reference-construction studies on matched subsets. The metric panel robustly distinguishes collaboration structures among systems with similar accuracy across methods and backbone models, while revealing framework-induced patterns that remain stable across different setups. We also obtain a series of substantive insights. Orchestration gains depend on backbone capability, and failures with similar outcomes can further reflect either structural mismatch or content-level error, implying different repairs. These results should be interpreted as descriptive evidence about observed outcomes and structural alignment. We argue that \ours~provides a consistent and structurally grounded basis for evaluating multi-agent methods.

We summarize our contributions as follows:
\begin{itemize}
\item We propose a graph-based evaluation framework that normalizes heterogeneous MAS execution traces into a unified graph representation, enabling horizontal evaluation across methods under the same structural criteria.
\item We instantiate the framework as \ours, releasing $844$ queries and their query-specific reference forests of verified-success collaboration graphs, together with adapters and a deterministic graph-metric panel.
\item We evaluate six MAS frameworks and use controlled backbone, reference-construction, perturbation, and failure diagnostic studies to characterize what the structural signals capture, where they are stable, and how they complement final-answer accuracy.
\end{itemize}
\section{Related Work}

\subsection{Multi-Agents Systems}

Recent work on LLM-based multi-agent systems has explored a wide spectrum of coordination paradigms. Early efforts focused on role-specialized pipelines, in which agents are assigned distinct responsibilities and communicate through predefined protocols~\citep{qian2024chatdev,dong2024self,talebirad2023multi}. Subsequent works introduce debate and critique mechanisms, where multiple agents iteratively argue, verify, or refine each other's intermediate answers to mitigate hallucination and single-agent bias~\citep{liang2024encouraging,khan2024debating,smit2023should}. More recently, researchers have turned to dynamic and self-organizing topologies, where the interaction graph itself is learned, searched, or adapted online to fit the task at hand~\citep{li2024improving,ishibashi2024self,zhang2024g}. Recent advances in MAS have therefore given rise to optimization directions. Adjacency matrices are dynamically optimized or the graph selector is explicitly parameterized to eliminate redundancy and adapt connectivity across rounds~\citep{wang2025agentdropout, li2025adaptive}. Another line of work goes beyond topology design and examines collaboration dynamics, ranging from causal analyses of sparsity’s impact on message propagation~\citep{shen2025understanding} to decentralized, reward or contribution driven self-organization~\citep{zhou2025reso}. Despite the diversity of MAS methods, most frameworks are still proposed and validated in relative isolation: each paper typically introduces a new topology or prompting scheme and reports final-task accuracy on a handful of datasets. This limitation has induced the development of evaluation protocols that go beyond the final answer to examine collaborative structures in depth and evaluate across methods.

\subsection{Agent Benchmarks and Evaluation}

Benchmarking and evaluation of LLM agents have developed rapidly. One strand builds task-oriented benchmarks that stress different domain-specific capabilities, such as deep research or analytical query\citep{liu2024agentbench, huang2026mmdeepresearch, wang2025fdabench}. Research into interactive and social contexts primarily investigates whether agents are capable of engaging in negotiation, collaboration, or competition within environments characterized by incomplete information and dynamically changing objectives\citep{xu2024magic, ma2024agentboard, wu2024smartplay}. In addition, many studies focus on process-level signal analysis of individual agents, primarily centering on their collaborative trajectories with specific metrics encompassing the fidelity of reasoning chains, the execution efficiency of task trajectories, and step-based reward modeling\citep{wang2026trajectory2task, wang2026aligning,chen2026contextual}. Other studies shift from outcome-only assessment to full-path evaluation, quantifying not just final performance but also behavioral legality, ordering, safety, and strategic competencies such as backtracking, decomposition, and self-verification~\citep{michelakis2025core,ou2025agentdiagnose}. However, existing step-level or trajectory-level measures are designed for one reasoning chain. A few recent works begin to probe MAS-specific failure modes or communication efficiency~\citep{zhang2024chain,pan2025agentcoord}, but a unified view that turns an entire MAS run into a comparable structural object is still missing. A graph-based perspective may offer a potential solution.

\begin{figure*}[t]
    \centering
    \includegraphics[width=1.0\textwidth]{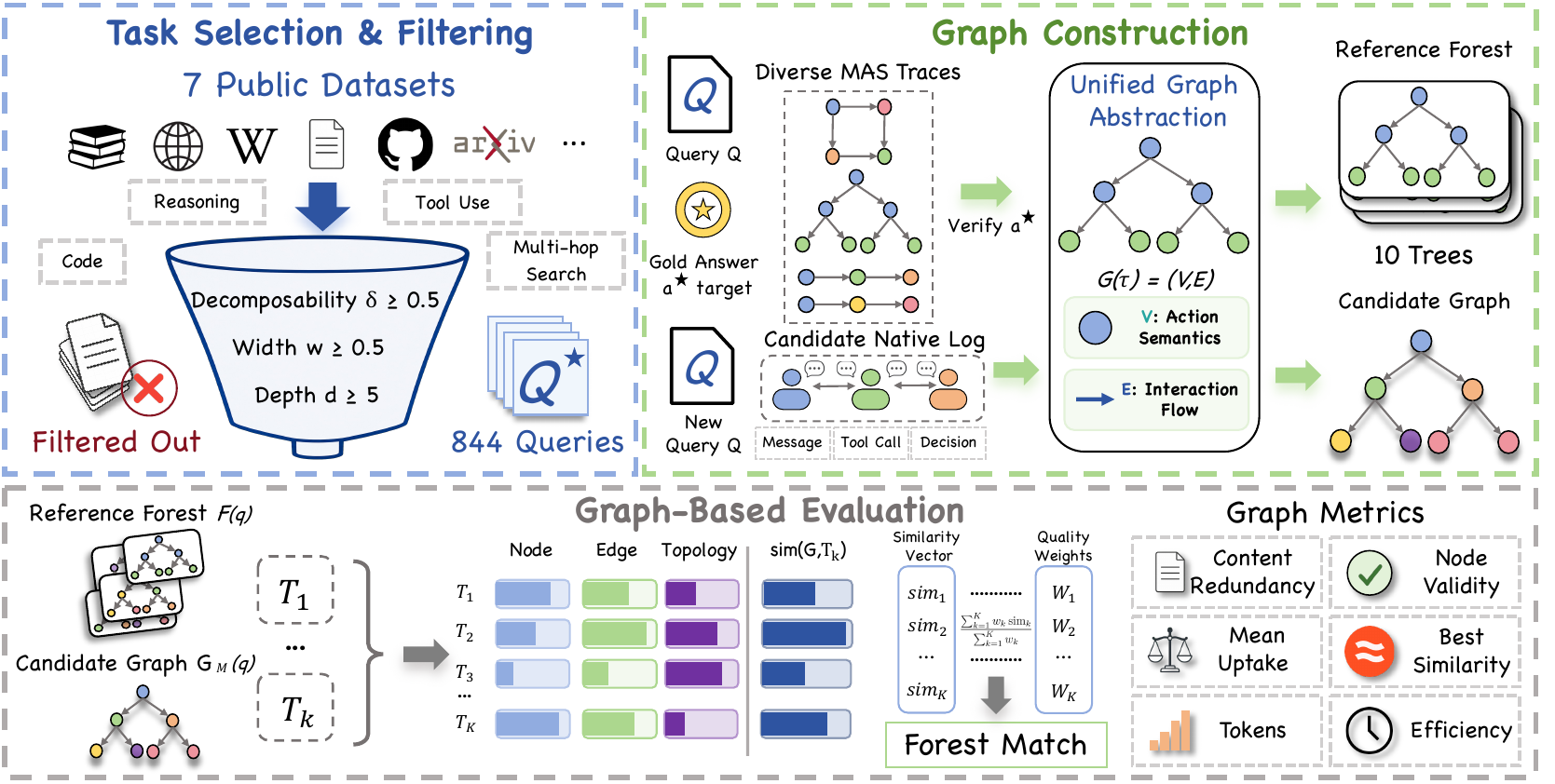}
    \caption{Overview of \ours~framework.}
    \label{fig:framework}
\end{figure*}

\section{Evaluation Framework}
\label{sec:framework}

\subsection{Graph Abstraction}
\label{sec:graph-abstraction}

A multi-agent system $\mathcal{M}$ executes a query $q$ by producing a temporally ordered trace $\tau=\mathcal{M}(q)$, namely the full sequence of messages, tool invocations, and decisions through which its agents jointly arrive at an answer. Because raw traces differ in shape across frameworks and are not directly comparable, we project every trace onto a uniform collaboration graph:
\begin{equation}
G(\tau)=(V,E,\phi),
\label{eq:exec-graph}
\end{equation}
where $V$ contains one node per atomic action of $\tau$, $E$ contains one directed edge per information transfer between two actions, and $\phi$ labels each node with its issuing agent, a canonical role $r\in\mathcal{R}$ drawn from a shared role vocabulary, and the textual content of the action. We do not split a single action into multiple nodes nor merge consecutive utterances by the same agent, so that the granularity of $G(\tau)$ matches the granularity of the trace itself.

This abstraction is well-defined as a directed acyclic graph because every atomic action is uniquely time-stamped at emission and an action can only consume information already emitted, so all edges point strictly forward in time and no cycle can form. The projection is intentionally lossy: it discards framework-specific runtime state and exact wall-clock interleavings, while preserving the roles, actions, and information-flow dependencies required by our structural metrics. Independent actions whose timestamps happen to interleave differently therefore collapse to the same partial order, and within-batch reorderings produce the same graph. This invariance is what makes native traces from otherwise incompatible frameworks structurally evaluable and comparable.

\subsection{Reference Forest}
\label{sec:reference-forest}

A task rarely admits a single successful collaboration pattern. The same task may be solved by debate, role-specialized agents, or a peer swarm. These executions can differ substantially in topology while producing the same verified answer, so using one process as ground truth would arbitrarily privilege one MAS paradigm.

\para{From a reference tree to a reference forest.} We regard each verified-success collaboration structure as a reference tree: it captures one way in which agent roles and information flows grow toward a correct outcome. Since several structurally distinct trees may be valid for the same query, its process-level reference should be a forest rather than one privileged tree. Operationally, a reference tree is represented by the collaboration DAG extracted from its execution trace; ``tree'' names its role in the forest metaphor, while the DAG is the object used for computation. We therefore define $\mathcal{F}(q)=\{(G_k,q_k)\}_{k=1}^{K(q)}$, where each $G_k$ is a verified-success reference graph sampled from a representative collaboration paradigm, and verified to complete the query correctly.

\para{Reference forest as a benchmark asset.} The forest is not a unique gold-standard process and does not enumerate every valid way to solve $q$. It is a reference set supplied by the benchmark: every included reference tree reaches the verified target outcome, and the forest covers multiple observed collaboration paradigms. The score $q_k$ records transparent within-forest preferences for correctness, structural parsimony, and token cost; it should not be interpreted as a universal process-quality label. This construction makes evaluation reusable: once the reference forest is released, a new trace can be compared with the same fixed collection of successful structures without regenerating the reference set.

Accordingly, our evaluation answers a scoped question: how does a candidate's observed collaboration structure relate to the benchmark's set of verified-success structures for the same query? It does not claim that structural similarity alone determines whether one collaboration is universally better than another. Outcome correctness remains separately measured by task accuracy.

\subsection{Evaluation Metric}
\label{sec:metric}

Once a trace has been reduced to its collaboration graph, evaluating it amounts to extracting structural and semantic statistics from the graph and, when a reference forest is available, comparing the graph against it. Because every quantity we report is a deterministic function of $G(\tau)$ alone, the evaluation needs no further LLM calls and runs in milliseconds per trace.

\para{Pairwise graph similarity.} Given a candidate graph $G$ and the reference graph $G'$ representing one tree in the forest, we measure their pairwise similarity as
\begin{equation}
\mathrm{sim}(G,G')=\alpha\,s_{\mathrm{node}}+\beta\,s_{\mathrm{edge}}+\gamma\,s_{\mathrm{topo}},
\label{eq:sim}
\end{equation}
with default weights $(\alpha,\beta,\gamma)=(0.4,0.4,0.2)$. Let $R(G)$ be the multiset of canonical node roles in $G$ and let $B(G)$ be the set of role-typed edge fingerprints $(r_u,r_v)$. We define
\begin{equation}
s_{\mathrm{node}}=\frac{|R(G)\cap R(G')|}{|R(G)\cup R(G')|},\quad
s_{\mathrm{edge}}=\frac{|B(G)\cap B(G')|}{|B(G)\cup B(G')|},
\end{equation}
where multiset intersection and union are used for $R(\cdot)$. The topology term compares graph size without rewarding verbosity:
\begin{equation}
s_{\mathrm{topo}}=\frac{1}{2}\left(\frac{\min(|V|,|V'|)}{\max(|V|,|V'|)}+\frac{\min(|E|,|E'|)}{\max(|E|,|E'|)}\right).
\end{equation}
Computing $s_{\mathrm{node}}$ and $s_{\mathrm{edge}}$ on canonicalized roles rather than raw agent identifiers makes $\mathrm{sim}$ invariant to agent renaming and to permutations of independent actions.

\para{Forest Match.} Each reference graph carries a continuous score $q_k>0$ combining verified correctness, structural parsimony, and token cost. We normalize $w_k=q_k/\max_{k'}q_{k'}\in(0,1]$ so that the highest-scoring reference receives weight $1$. Forest Match aggregates pairwise similarities into a weighted average:
\begin{equation}
\mathrm{FM}(G,\mathcal{F}(q))=\frac{\sum_{k=1}^{K(q)} w_k\,\mathrm{sim}(G,G_k)}{\sum_{k=1}^{K(q)} w_k}\;\in\;[0,1].
\label{eq:forest-match}
\end{equation}
Forest Match measures how broadly a candidate aligns with the successful structures represented across the forest. Because a candidate may closely resemble one valid reference tree without matching the others, we also report \textit{Best Similarity}, $\max_k\mathrm{sim}(G,G_k)$. Forest Match captures forest-level alignment, whereas Best Similarity captures proximity to the nearest observed reference tree; neither is interpreted as a stand-alone measure of universal collaboration quality.

\para{Auxiliary metrics.} Beyond \textit{Accuracy}, the graph exposes a compact metric panel. Let $V^{-}\subseteq V$ denote actions eligible for downstream consumption, excluding designated terminal-output nodes. \textit{Node Validity} measures the fraction of these actions consumed by at least one downstream action,
\begin{equation}
\mathrm{NV}(G)=\frac{|\{v\in V^{-}: \exists (v,u)\in E\}|}{|V^{-}|}.
\end{equation}
\textit{Mean Uptake} measures whether downstream content semantically uses upstream content,
\begin{equation}
\mathrm{Uptake}(G)=\frac{1}{|E|}\sum_{(u,v)\in E}\mathbf{1}[\cos(\mathrm{emb}(x_u),\mathrm{emb}(x_v))>\theta_u],
\end{equation}
where $x_v$ is the textual content attached to node $v$ and $\theta_u$ is fixed across all methods. \textit{Content Redundancy} measures duplicate content after normalization,
\begin{equation}
\mathrm{CR}(G)=1-\frac{|\mathrm{unique}(\{h(x_v):v\in V\})|}{|V|},
\end{equation}
where $h(\cdot)$ is the normalized content hash. \textit{Tokens} is the total backbone-token budget consumed by the trace, and \textit{Efficiency} is topology efficiency, computed as graph parallelism per active agent. Read together with accuracy, Forest Match, and Best Similarity, these quantities profile a candidate along outcome, reference alignment, content flow, and budget. They expose ignored or repeated work as observable trace properties without assigning a universal quality label to the entire collaboration.

\section{The \ours~Benchmark}
\label{sec:benchmark}

We instantiate the framework above as a concrete benchmark with three released components: a fixed set of queries selected for structural analysis, a query-specific reference forest of verified-success graphs, and a uniform evaluation interface that converts candidate MAS traces into the shared representation.

\subsection{Data Collection}
\label{sec:task-data}

\ours{} is built from a curated pool of existing benchmarks that cover a broad spectrum of task types and capability demands.  
The raw query set $\mathcal{Q}_0$ is drawn from seven publicly available datasets, selected to represent four major task categories:

\begin{itemize}
    \item \textbf{Reasoning}: GPQA, MMLU, MATH-500, GSM8K
    \item \textbf{Code generation}: SWE-bench-Verified
    \item \textbf{Tool use}: Terminal-Bench
    \item \textbf{Multi-hop search}: WideSearch
\end{itemize}

\subsection{Query Filtering}
\label{sec:task-filter}

A persistent pitfall in MAS evaluation is that many standard items are too short, narrow, or serial to expose meaningful structural differences among collaboration methods. On such items, a well-prompted single agent with a highly capable LLM backbone may already be sufficient, and additional agents mainly add calls rather than an informative collaboration trace.

To focus the benchmark on queries more likely to elicit non-trivial, comparable collaboration structures, \ours{} applies a Trace-Suitability Pipeline (TSP). Each query $q \in \mathcal{Q}_0$ is independently scored by a strong filtering LLM along three dimensions:

\begin{itemize}
    \item \textbf{Depth $d(q) \in \mathbb{Z}_{\ge 1}$}: the minimum number of distinct reasoning steps required, suppressing trivial factual lookups.
    \item \textbf{Width $w(q) \in [0,1]$}: the diversity of required capabilities, computed as the normalized Shannon entropy over scores assigned to eight competencies (factual recall, logical reasoning, mathematics, coding, search, creative writing, domain knowledge, and planning).
    \item \textbf{Decomposability $\delta(q) \in [-2,+2]$}: whether the query admits a natural partition into independent sub-tasks, suppressing intrinsically serial problems.
\end{itemize}

A query is retained only if all three criteria are simultaneously satisfied:
\begin{equation}
d(q) \ge \theta_d,\quad
w(q) \ge \theta_w,\quad
\delta(q) \ge \theta_\delta,
\end{equation}
with thresholds $(\theta_d, \theta_w, \theta_\delta) = (5,\, 0.5,\, 0.5)$.
The conjunction-based design excludes a query when any one structural prerequisite is near-trivial. TSP is therefore a heuristic task-selection mechanism for eliciting analyzable collaboration traces; it does not by itself prove that every retained item requires multiple agents or that a single agent cannot solve it. Hyperparameter analysis of the thresholds can be found in \secref{app:supplementary-validation}.

\begin{figure}[t]
\centering
\includegraphics[width=1\linewidth]{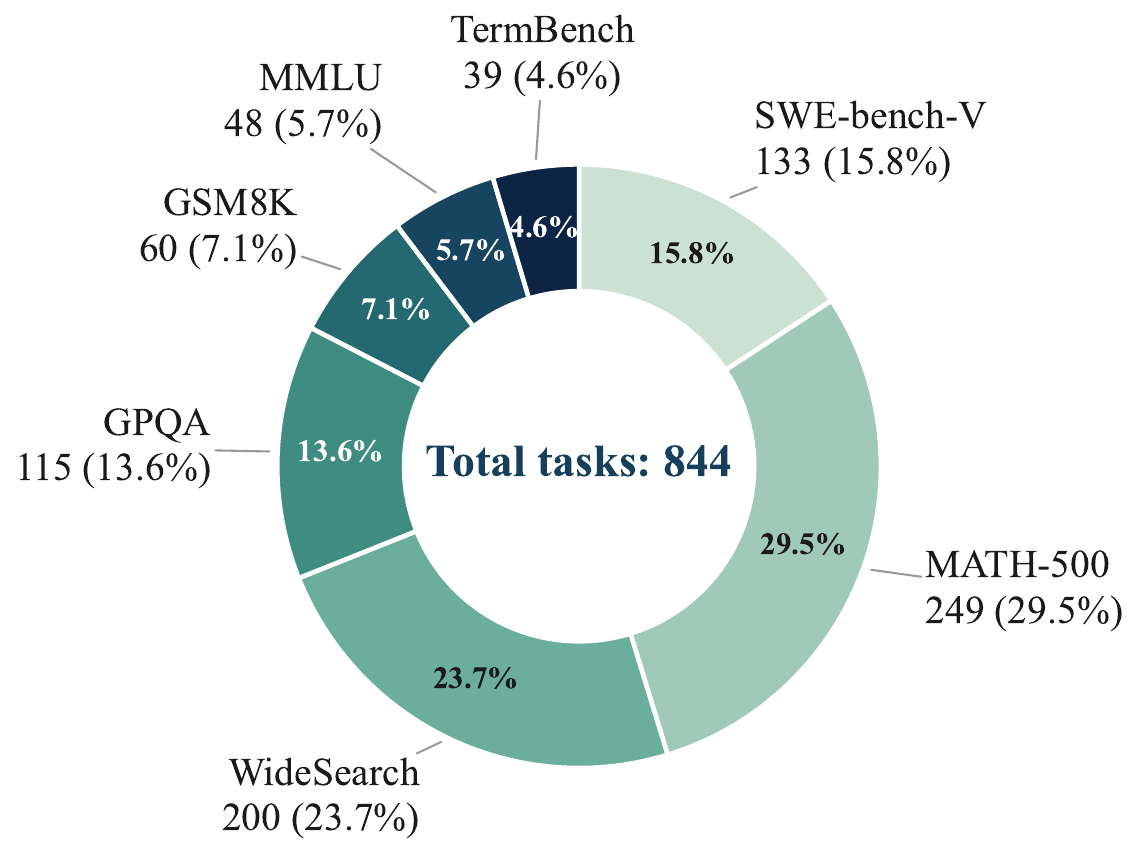}
\caption{Per-source composition of \ours{} after Trace-Suitability filtering.}
\label{fig:placeholder}
\end{figure}

After filtering, the final query set is:
\begin{equation}
\mathcal{Q}_\star = \{ q \in \mathcal{Q}_0 : \mathrm{TSP}(q) = 1 \}
\end{equation}
which contains $844$ queries distributed across reasoning ($55.9\%$), multi-hop QA ($23.7\%$), code generation ($15.8\%$), and tool use ($4.6\%$). The per-source composition of $\mathcal{Q}_\star$ is shown in Figure~\ref{fig:placeholder}.

\subsection{Reference Forest Construction}
\label{sec:reference-forest-construction}

For every $q\in\mathcal{Q}_\star$ with verified answer $a^\star$, we construct $\mathcal{F}(q)$ by sampling successful traces and projecting them through Section~\ref{sec:graph-abstraction}. To cover diverse observed structures, the reference generators include six representative MAS frameworks: Multi-agent Debate\cite{du2023improving}, Swarm\cite{openai2024swarm}, AutoGen\cite{wu2024autogen}, MetaGPT\cite{hong2024metagpt}, MAS-GPT\cite{ye2025mas}, and AFlow\cite{zhang2025aflow}. We sample each generator multiple times and retain $10$ verified-success trajectories per query, sampled across multiple methods. In the main construction, $a^\star$ is provided as a target so that generation focuses on producing candidate processes that reach the known-correct outcome. This conditioning is a reference-construction device, not evidence that the resulting trace is the unique or optimal process. Each generated trace must reproduce $a^\star$ before inclusion, after which it is clustered by structural skeleton and assigned a reference weight combining verified correctness, structural parsimony, and token cost. The procedure yields $K=10$ reference graphs per query and $8{,}440$ in total.
\subsection{Evaluating a New Candidate}
\label{sec:graph-construction}

Once $\mathcal{Q}_\star$ and its reference forests are fixed, scoring a new candidate MAS on a query $q\in\mathcal{Q}_\star$ proceeds in three deterministic steps. We first run the candidate on $q$ and parse its native trajectory into a unified event stream of three kinds, \texttt{agent\_message}, \texttt{tool\_call}, and \texttt{decision}, with one short conversion routine per framework so that no change is required in the underlying system. The stream is then projected to a collaboration graph $G_{\mathcal{M}}(q)$, and raw agent names are mapped into the shared role vocabulary $\mathcal{R}$ through a per-framework alias table that lets semantically identical roles across frameworks be compared on equal footing. Finally, the candidate is evaluated against the precomputed $\mathcal{F}(q)$ using the evaluation metrics defined in \secref{sec:metric}.

\section{Experiments}
In this section, we conduct comprehensive experiments to answer the following research questions. 
\begin{itemize}
    \item \textbf{RQ1 (Horizontal Evaluation):} Can a shared graph representation expose comparable collaboration differences across heterogeneous MAS methods?
    \item \textbf{RQ2 (Validity and Robustness):} How sensitive are the comparisons to the backbone, generator, and answer conditioning used to construct the reference forest?
    \item \textbf{RQ3 (Metric Behavior):} Do the graph metrics respond predictably to controlled structural changes?
    \item \textbf{RQ4 (Failure Diagnosis):} What actionable failure patterns become visible when structural alignment is analyzed together with outcome correctness?
    \item \textbf{RQ5 (Efficiency):} What are the online and amortized costs relative to LLM-as-Judge evaluation?
\end{itemize}
The main text reports the headline comparisons, reference validity, controlled metric behavior, coarse failure diagnostics, and evaluation cost. Complete perturbation, sensitivity, backbone, source-level, and diagnostic details are provided in the appendix.
\subsection{Implementation Details}

\paragraph{Candidate frameworks.} We evaluate six representative MAS frameworks that span the design spectrum of recent literature: \textbf{Multi-agent Debate}\cite{du2023improving}, \textbf{Swarm}\cite{openai2024swarm}, \textbf{AutoGen}\cite{wu2024autogen}, \textbf{MetaGPT}\cite{hong2024metagpt}, \textbf{MAS-GPT}\cite{ye2025mas}, and \textbf{AFlow}\cite{zhang2025aflow}.

\paragraph{Backbone.} We use DeepSeek-V4-Flash\cite{deepseekai2026deepseekv4} as the default candidate backbone and to construct the default reference forest. In subsequent comparison and evaluation experiments, we additionally use Claude-Sonnet-4.6\cite{anthropic2026sonnet46}, GPT-5.5\cite{openai2026introducing}, and Qwen3.6-Plus\cite{qwen36plus}.

\paragraph{Metrics.} We evaluate the MAS methods using the metrics defined in \secref{sec:metric}. We apply Leave-One-Out (LOO) processing when candidate and reference data share a framework source, so that structural evaluation does not directly compare a candidate with references generated by the same framework.

\begin{table*}[t]
\centering
\small
\caption{Global performance on \ours~with the diverse DeepSeek-V4-Flash reference forest. Best per column is in \textbf{bold}; lower is better for Content Redundancy and Tokens. Acc. denotes task accuracy and Eff. denotes topology efficiency, computed as parallelism per agent in the collaboration graph.}
\label{tab:main}
\resizebox{\linewidth}{!}{
\begin{tabular}{lcccccccc}
\toprule
Framework & Acc.\,$\uparrow$ & Forest Match\,$\uparrow$ & Node Validity\,$\uparrow$ & Mean Uptake\,$\uparrow$ & Content Redundancy\,$\downarrow$ & Best Similarity\,$\uparrow$ & Tokens\,$\downarrow$ & Eff.\,$\uparrow$ \\
\midrule
MAS-GPT & 0.546 & \textbf{0.421} & \textbf{1.000} & 0.267 & 0.009 & \textbf{0.775} & 2{,}457 & 0.830 \\
AutoGen & 0.551 & 0.406 & \textbf{1.000} & 0.210 & 0.011 & 0.579 & \textbf{1{,}783} & 1.998 \\
MetaGPT & 0.533 & 0.395 & \textbf{1.000} & \textbf{0.324} & 0.022 & 0.559 & 2{,}329 & 0.833 \\
Swarm   & 0.531 & 0.350 & 0.968 & 0.253 & 0.051 & 0.462 & 1{,}936 & 1.096 \\
AFlow   & 0.538 & 0.347 & \textbf{1.000} & 0.309 & 0.167 & 0.507 & 2{,}432 & 0.844 \\
Debate  & \textbf{0.569} & 0.316 & \textbf{1.000} & 0.249 & \textbf{0.000} & 0.284 & 3{,}227 & \textbf{2.258} \\
\bottomrule
\end{tabular}
}
\end{table*}

\subsection{Performance of Different MAS Methods}
\label{sec:exp-main}

In this section, we conduct a comparative evaluation of \ours~ across different frameworks. Each framework is run with 3 random seeds per query, yielding roughly $3 \times 844 \approx 2{,}532$ traces per framework and $\sim$15{,}192 traces in total. Table~\ref{tab:main} reports the global performance of the six candidate frameworks under the diverse DeepSeek-V4-Flash reference forest. All columns follow the metric panel of Section~\ref{sec:metric}. We propose the following observations.

\vpara{Accuracy alone weakly separates the evaluated methods.} The six candidate accuracies span $0.038$, from $0.531$ to $0.569$. Debate has the highest accuracy but the lowest Forest Match and Best Similarity, while also consuming nearly twice the tokens of the cheapest method. At the same time, Debate has the lowest content redundancy and the highest topology efficiency. These differences illustrate why no single scalar captures the whole collaboration trace: outcome, reference alignment, content reuse, and cost must be read as a panel.

\vpara{The forest provides a discriminative structural signal.} Forest Match ranges from $0.316$ (Debate) to $0.421$ (MAS-GPT), whereas the corresponding accuracies are tightly clustered. Forest Match summarizes alignment across multiple reference trees, while Best Similarity asks whether a candidate closely matches at least one tree in the forest. Neither asserts that more elaborate orchestration is better. The ordering can therefore change by domain: on SWE-bench-Verified, whose forest contains more review-and-revise reference trees, Debate reaches $0.382$ and becomes the most aligned method. The forest thus reveals which collaboration shapes are broadly or selectively compatible with successful patterns for a task, complementing rather than replacing accuracy.

\vpara{The structural panel exposes differences hidden by the final answer.} The metrics separate methods with comparable accuracy by surfacing distinct trace properties. Debate's Best Similarity is $0.284$, the lowest of the six, indicating limited proximity to any one observed reference mode despite its low redundancy and high topology efficiency. AFlow's content redundancy reaches $0.167$, roughly three times the next framework, reflecting repeated content across its parallel solvers. Swarm is the only framework with reported node validity below one at $0.968$, indicating that some actions are not consumed downstream under our extraction rule. None of these properties is visible from accuracy alone; together they support that \ours~makes heterogeneous collaboration structures directly comparable and diagnostically separable.

\subsection{Validity and robustness of the Reference Forest}
\label{sec:exp-reference-forest-robustness}
\label{sec:exp-perturb}
\label{sec:exp-reference-forest-ablation}

A natural concern about any reference-based metric is that its comparison could encode artifacts of how the reference forest was built. We therefore vary three construction choices that are central to our setup: reference backbone, reference generator, and gold-answer conditioning. First, we independently construct reference forests with DeepSeek-V4-Flash, Claude-Sonnet-4.6, and GPT-5.5. The six candidate methods retain the same aggregate ordering under all three forests, with pairwise Kendall $\tau$ and Spearman $\rho$ equal to $1.00$. This supports aggregate stability across the tested backbones, without claiming invariance to arbitrary reference construction.

\vpara{Reference-generator and conditioning robustness.} We first remove all reference traces authored by one MAS method at a time, and then rebuild the reference forest on a 30-task stratified subset without gold-answer conditioning.

The seven settings hold out each of the six MAS reference generators and the CoT-SC structural comparison in turn. This test is stricter than evaluation-time leave-one-out because it removes an entire contributor's reference traces. As shown in \tableref{tab:heldout-reference-forest}, the ranking remains positively aligned with the full reference ranking and preserves the Top-3 in 6/7 settings. This supports the view of a reference set of non-unique solutions presented in \secref{sec:reference-forest}. Appendix~\ref{app:reference-forest-robustness} reports the complementary leave-source-out ablation.

\begin{table}[t]
\centering
\caption{Held-out reference-generator ablation. Each setting removes all reference traces contributed by one of the six MAS frameworks or the CoT-SC anchor.}
\label{tab:heldout-reference-forest}
\begin{tabular}{lc}
\toprule
Result & Value \\
\midrule
Held-out settings & 7 \\
Kendall $\tau$ range & $0.60$--$1.00$ \\
Top-3 preserved & 6/7 \\
Lowest $\tau$ & $0.60$ \\
\bottomrule
\end{tabular}
\end{table}

\begin{table}[t]
\centering
\small
\caption{Gold-conditioned vs. open-ended reference-forest construction on a 30-task stratified validation subset. Values are mean Forest Match under each construction.}
\label{tab:openended-reference-forest}
\begin{tabular}{lccc}
\toprule
Framework & Gold-conditioned & Open-ended & $\Delta$ \\
\midrule
MAS-GPT & 0.4422 & 0.4914 & +0.049 \\
AutoGen & 0.4288 & 0.4691 & +0.040 \\
MetaGPT & 0.4097 & 0.4403 & +0.031 \\
Swarm & 0.3758 & 0.3919 & +0.016 \\
AFlow & 0.3681 & 0.4012 & +0.033 \\
Debate & 0.2999 & 0.2720 & -0.028 \\
\midrule
Ranking agreement & \multicolumn{3}{c}{Kendall $\tau=0.867$; Top-3 exactly preserved} \\
\bottomrule
\end{tabular}
\vspace{-15pt}
\end{table}

Table~\ref{tab:openended-reference-forest} shows that removing gold-answer conditioning changes the absolute Forest Match values, but largely preserves the aggregate ranking: MAS-GPT, AutoGen, and MetaGPT remain the Top-3, and only the middle ranks swap. We therefore treat the reference forest as a stable aggregate basis for trajectory-level evaluation and comparison, while avoiding claims that individual trace-level scores are invariant to every construction choice.

\subsection{Controlled Metric Behavior and Hyperparameter Sensitivity}
\label{sec:metric-behavior}

We apply node deletion, edge rewiring, role swaps, distractor insertion, and duplicate-content injection to a stratified trace sample. Structural damage consistently reduces Forest Match. Node deletion at $p=0.5$ produces the largest decrease of $0.058$. In contrast, duplicate-content injection leaves the set-based structural score nearly unchanged and is detected by Content Redundancy. These results illustrate the intended division of labor within the metric panel: Forest Match captures reference-relative structure, whereas auxiliary metrics capture properties such as repetition and uptake.

\vpara{The conclusions are stable under non-degenerate metric weights.} We compare the default weighting with uniform, node-heavy, edge-heavy, and topology-heavy settings. Across these five non-degenerate schemes, the framework ordering retains Kendall $\tau\ge0.73$ and preserves the same Top-3. The uniform, node-heavy, and edge-heavy schemes each yield Kendall $\tau\ge0.87$. The topology-only setting changes the ordering substantially, yielding $\tau=0.20$. This result is expected because the topology-only setting removes role and communication semantics from Eq.~\eqref{eq:sim}.

\vpara{The framework ranking is insensitive to the selected TSP thresholds.} We sweep all $36$ combinations of $\theta_d\in\{4,5,6\}$, $\theta_w\in\{0.4,0.5,0.6\}$, and $\theta_\delta\in\{-0.5,0.0,0.5,1.0\}$. These settings produce pools ranging from $222$ to $1{,}244$ queries. The framework ranking remains unchanged across all settings, with Kendall $\tau=1.00$ in every case. The Top-1 framework is preserved in all $36$ settings. These results show that the main comparison does not depend on default thresholds of $\theta_d=5$, $\theta_w=0.5$, and $\theta_\delta=0.5$. Appendix~\ref{app:supplementary-validation} reports the complete perturbation and sensitivity tables.

\subsection{Failure Modes Exposed by \ours{}}

\begin{figure*}[t]
\centering
\includegraphics[width=0.92\textwidth]{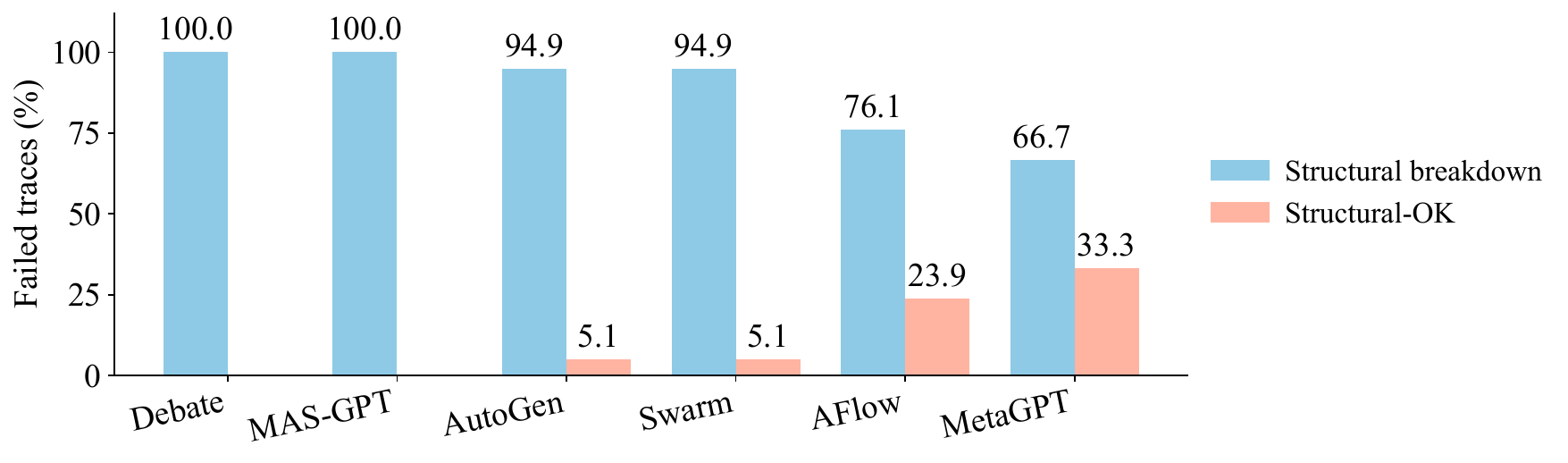}
\caption{Failure-mode diagnostics across the six candidate frameworks on failed traces. Structural-OK denotes traces whose collaboration shape remains plausible despite an incorrect answer; structural breakdown denotes traces with at least one non-OK structural label. The framework-level differences illustrate how the benchmark supports diagnosis.}
\label{fig:failure-modes}
\end{figure*}

In this section, we analyze the failure modes that \ours~revealed during actual experiments. A structural benchmark should not only separate methods globally, but also reveal why similarly scored outcomes arise from different collaboration processes. We therefore sample $300$ candidate traces stratified by framework, dataset, and the accuracy--Forest Match quadrant, and use an LLM to explain structural and content failures in light of the metrics. These labels are not used as benchmark scores; they are an interpretive layer over the deterministic graph metrics. Figure~\ref{fig:failure-modes} summarizes the resulting framework-level patterns on failed traces.

\vpara{Failures separate into structural and content mechanisms.} Some traces fail because the collaboration process is underspecified for the task: a useful role is missing or a framework repeats a fixed motif that does not fit the query. Other traces exhibit a plausible collaboration structure but still fail because a solver, verifier, or aggregator introduces or accepts incorrect content. The first mechanism motivates task-adaptive role and topology selection; the second motivates stronger verification, aggregation, and evidence uptake inside an otherwise reasonable process.

\vpara{Coarse diagnostic directions are stable across judges.} To reduce dependence on one LLM, we ran the same evaluation process using DeepSeek-V4-Flash, GPT-5.5, and Qwen3.6-Plus (which used as the judge backbone only) on the same workflow and aggregated labels by majority vote. Table~\ref{tab:judge-panel} shows high majority agreement on the shared parse-clean subset, supporting the use of broad failure axes for framework debugging. Fine-grained label frequencies of LLM--as--judge remain rubric-sensitive; the complete prompts and sensitivity results are reported in Appendix~\ref{app:diagnostic-scope}.

\begin{table}[t]
\centering
\small
\caption{Agreement of a three-model diagnostic panel on the shared parse-clean subset ($n=138$).}
\label{tab:judge-panel}
\begin{tabular}{lc}
\toprule
Diagnostic axis & $\ge2/3$ judges agree \\
\midrule
Structural failure & 93.5\% \\
Content failure & 81.2\% \\
Whether the error was caught & 100.0\% \\
First error location & 94.9\% \\
\bottomrule
\end{tabular}
\vspace{-15pt}
\end{table}

As a result, \ours~illustrates a practical use of \ours~beyond global ranking across methods: Forest Match localizes whether a failed run is structurally unusual relative to successful references, while accuracy and content diagnostics determine whether the remaining problem lies in reasoning or verification.

\subsection{Cost Analysis: \ours~vs.\ LLM-as-Judge}
\label{sec:exp-cost}

We compare the evaluation cost of \ours~against an LLM-as-Judge that reads the same traces and scores their apparent collaboration organization, communication efficiency, and role usage. We likewise use DeepSeek-V4-Flash as the backbone for evaluation.

\begin{figure}
    \centering
    \includegraphics[width=1\linewidth]{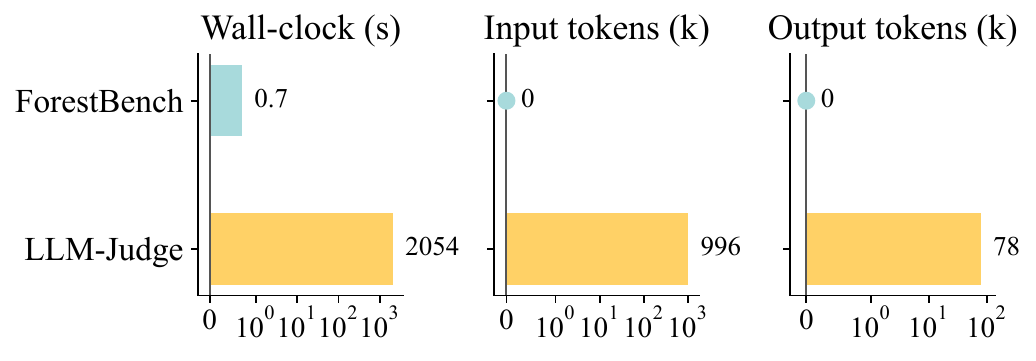}
    \caption{Online evaluation cost of \ours{} versus LLM-as-Judge for $1{,}000$ traces. \ours~is $\approx$ 2,900× faster and uses zero inference tokens per 1k traces.}
    \label{fig:cost}
\end{figure}

\vpara{Trace evaluation cost.} \ours~scores a candidate by graph matching against a pre-built reference forest, so each trace is evaluated with $0$ inference tokens and in milliseconds, against the multi-thousand judge-input tokens and seconds-of-inference each trace requires under LLM-as-Judge. On $1{,}000$ traces \ours~finishes in $0.7$\,s, while the LLM-Judge baseline takes roughly $34$ minutes; the gap widens linearly as the benchmark is rerun or extended, since \ours~pays no token cost per evaluation while LLM-Judge pays the same per-trace token cost on every run.

\vpara{One-off reference-forest construction cost.} We additionally report the one-off cost of building the reference forest. DeepSeek-V4-Flash consumes $20.5$M tokens to construct over the $844$-query pool, averaging $24.3$k tokens per query across $8{,}440$ target-conditioned reference traces (in the same accounting unit as the Judge column above). This is a non-trivial cost but paid once: the resulting forest evaluates every candidate framework, every backbone swap, and every robustness experiment in this paper with no additional inference time or tokens. We release the reference forests alongside the benchmark so that downstream users inherit it without paying the construction cost again, and extending the pipeline to a new dataset incurs this cost once and zero per evaluation thereafter.

\section{Discussion}
\label{sec:insights}

\vpara{A common graph space enables horizontal evaluation.} The primary contribution of \ours{} is a shared representation for comparing heterogeneous methods, rather than a new free-form judgment of collaboration quality. Native logs differ in event schemas, role names, and communication protocols; graph abstraction maps them to a common metric panel covering outcome, reference alignment, information uptake, redundancy, topology, and cost. The narrow $0.038$ accuracy range in Table~\ref{tab:main} contrasts with wider structural variation. Stability across balanced Forest Match weights and all $36$ TSP-threshold settings further shows that the comparison is not driven by one hyperparameter configuration.

\vpara{The forest replaces single-template matching with plural comparison.} A single reference tree would tie process evaluation to one coordination template. The released forest retains multiple verified-success structures for each query and compares every candidate against the same reference set. Cross-backbone, held-out-generator, open-ended, and leave-source-out tests preserve the main aggregate comparisons despite changes in Forest Match values and some middle ranks. Forest Match should therefore be interpreted as reference-relative alignment across the forest, while Best Similarity measures proximity to the nearest reference tree.

\vpara{Framework and backbone effects can be separated under the shared representation.} Appendix~\ref{app:backbone-analysis} indicates that frameworks impose recognizable structural priors, whereas the accuracy benefit of orchestration depends on the backbone. Relative to a single-agent CoT baseline, mean orchestration gains are $-0.002$ for DeepSeek-V4, $-0.039$ for GPT-5.5, and $-0.069$ for Claude-S4.6. Adding agents is therefore not uniformly beneficial; the relevant question is whether a collaboration pattern complements the task.

\vpara{Aggregate alignment can hide domain specialization.} Debate has the lowest aggregate Forest Match but ranks first on SWE-bench-V, where review-and-revise behavior is more common in successful reference graphs. Global rankings should therefore be complemented with source-level analysis for domain-specific method selection. Different methods may be better suited to specific tasks.

\section{Conclusion}

In this paper, we presented a unified graph abstraction approach for evaluating and comparing heterogeneous multi-agent execution traces. The framework maps native multi-agent collaborative trajectories into directed acyclic collaboration graphs and evaluates each candidate against a query-specific forest of benchmark-provided, verified-success reference graphs. We instantiated the framework as \ours, releasing collaboration-necessary queries, their fixed reference forests, adapters, and a deterministic graph-metric panel. Across \ours~and six representative MAS methods, with controlled backbone, reference-forest-construction, and perturbation studies, we show that \ours~(i) separates collaboration structures that final-answer accuracy leaves indistinguishable, (ii) exposes framework- and domain-specific collaboration motifs, (iii) provides stable signals under the tested construction and perturbation settings, and (iv) evaluates new traces orders of magnitude faster than repeatedly invoking an LLM judge. By abstracting the collaborative trajectories of heterogeneous frameworks into a unified graph structure, our approach offers a new perspective on MAS evaluation, enables more consistent and comparable evaluation.

\bibliographystyle{ACM-Reference-Format}
\bibliography{custom}

@article{li2024survey,
  title={A survey on LLM-based multi-agent systems: workflow, infrastructure, and challenges},
  author={Li, Xinyi and Wang, Sai and Zeng, Siqi and Wu, Yu and Yang, Yi},
  journal={Vicinagearth},
  volume={1},
  number={1},
  pages={9},
  year={2024},
  publisher={Springer}
}

@article{han2024llm,
  title={LLM multi-agent systems: Challenges and open problems},
  author={Han, Shanshan and Zhang, Qifan and Jin, Weizhao and Xu, Zhaozhuo},
  journal={arXiv preprint arXiv:2402.03578},
  year={2024}
}

@article{guo2024large,
  title={Large language model based multi-agents: A survey of progress and challenges},
  author={Guo, Taicheng and Chen, Xiuying and Wang, Yaqi and Chang, Ruidi and Pei, Shichao and Chawla, Nitesh V and Wiest, Olaf and Zhang, Xiangliang},
  journal={arXiv preprint arXiv:2402.01680},
  year={2024}
}

@inproceedings{wu2024autogen,
  title={Autogen: Enabling next-gen LLM applications via multi-agent conversations},
  author={Wu, Qingyun and Bansal, Gagan and Zhang, Jieyu and Wu, Yiran and Li, Beibin and Zhu, Erkang and Jiang, Li and Zhang, Xiaoyun and Zhang, Shaokun and Liu, Jiale and others},
  booktitle={First conference on language modeling},
  year={2024}
}

@inproceedings{hong2024metagpt,
  title={MetaGPT: Meta programming for a multi-agent collaborative framework},
  author={Hong, Sirui and Zhuge, Mingchen and Chen, Jonathan and Zheng, Xiawu and Cheng, Yuheng and Wang, Jinlin and Zhang, Ceyao and Yau, Steven and Lin, Zijuan and Zhou, Liyang and others},
  booktitle={International Conference on Learning Representations},
  volume={2024},
  pages={23247--23275},
  year={2024}
}

@inproceedings{chen2024agentverse,
  title={Agentverse: Facilitating multi-agent collaboration and exploring emergent behaviors},
  author={Chen, Weize and Su, Yusheng and Zuo, Jingwei and Yang, Cheng and Yuan, Chenfei and Chan, Chi-Min and Yu, Heyang and Lu, Yaxi and Hung, Yi-Hsin and Qian, Chen and others},
  booktitle={International Conference on Learning Representations},
  volume={2024},
  pages={20094--20136},
  year={2024}
}

@inproceedings{zhuge2024gptswarm,
  title={Gptswarm: Language agents as optimizable graphs},
  author={Zhuge, Mingchen and Wang, Wenyi and Kirsch, Louis and Faccio, Francesco and Khizbullin, Dmitrii and Schmidhuber, J{\"u}rgen},
  booktitle={Forty-first International Conference on Machine Learning},
  year={2024}
}

@article{cemri2026multi,
  title={Why do multi-agent llm systems fail?},
  author={Cemri, Mert and Pan, Melissa Z and Yang, Shuyi and Agrawal, Lakshya A and Chopra, Bhavya and Tiwari, Rishabh and Keutzer, Kurt and Parameswaran, Aditya and Klein, Dan and Ramchandran, Kannan and others},
  journal={Advances in Neural Information Processing Systems},
  volume={38},
  year={2026}
}

@article{cobbe2021training,
  title={Training verifiers to solve math word problems},
  author={Cobbe, Karl and Kosaraju, Vineet and Bavarian, Mohammad and Chen, Mark and Jun, Heewoo and Kaiser, Lukasz and Plappert, Matthias and Tworek, Jerry and Hilton, Jacob and Nakano, Reiichiro and others},
  journal={arXiv preprint arXiv:2110.14168},
  year={2021}
}

@article{hendrycks2020measuring,
  title={Measuring massive multitask language understanding},
  author={Hendrycks, Dan and Burns, Collin and Basart, Steven and Zou, Andy and Mazeika, Mantas and Song, Dawn and Steinhardt, Jacob},
  journal={arXiv preprint arXiv:2009.03300},
  year={2020}
}

@article{chen2021evaluating,
  title={Evaluating large language models trained on code},
  author={Chen, Mark and Tworek, Jerry and Jun, Heewoo and Yuan, Qiming and Pinto, Henrique Ponde De Oliveira and Kaplan, Jared and Edwards, Harri and Burda, Yuri and Joseph, Nicholas and Brockman, Greg and others},
  journal={arXiv preprint arXiv:2107.03374},
  year={2021}
}

@inproceedings{chan2024chateval,
  title={Chateval: Towards better llm-based evaluators through multi-agent debate},
  author={Chan, Chi-Min and Chen, Weize and Su, Yusheng and Yu, Jianxuan and Xue, Wei and Zhang, Shanghang and Fu, Jie and Liu, Zhiyuan},
  booktitle={International conference on learning representations},
  volume={2024},
  pages={9079--9093},
  year={2024}
}

@article{chen2026contextual,
  title={Contextual Counterfactual Credit Assignment for Multi-Agent Reinforcement Learning in LLM Collaboration},
  author={Chen, Yanjun and Sun, Yirong and Wang, Hanlin and Zhang, Xinming and Shen, Xiaoyu and Li, Wenjie and Zhang, Wei},
  journal={arXiv preprint arXiv:2603.06859},
  year={2026}
}

@article{yehudai2025survey,
  title={Survey on evaluation of llm-based agents},
  author={Yehudai, Asaf and Eden, Lilach and Li, Alan and Uziel, Guy and Zhao, Yilun and Bar-Haim, Roy and Cohan, Arman and Shmueli-Scheuer, Michal},
  journal={arXiv preprint arXiv:2503.16416},
  year={2025}
}

@inproceedings{zhou2025reso,
  title={Reso: A reward-driven self-organizing llm-based multi-agent system for reasoning tasks},
  author={Zhou, Heng and Geng, Hejia and Xue, Xiangyuan and Kang, Li and Qin, Yiran and Wang, Zhiyong and Yin, Zhenfei and Bai, Lei},
  booktitle={Proceedings of the 2025 Conference on Empirical Methods in Natural Language Processing},
  pages={15990--16009},
  year={2025}
}

@inproceedings{zhang2025aflow,
  title={Aflow: Automating agentic workflow generation},
  author={Zhang, Jiayi and Xiang, Jinyu and Yu, Zhaoyang and Teng, Fengwei and Chen, Xionghui and Chen, Jiaqi and Zhuge, Mingchen and Cheng, Xin and Hong, Sirui and Wang, Jinlin and others},
  booktitle={International Conference on Learning Representations},
  volume={2025},
  pages={34040--34077},
  year={2025}
}

@inproceedings{qian2025scaling,
  title={Scaling large language model-based multi-agent collaboration},
  author={Qian, Chen and Xie, Zihao and Wang, Yifei and Liu, Wei and Zhu, Kunlun and Xia, Hanchen and Dang, Yufan and Du, Zhuoyun and Chen, Weize and Yang, Cheng and others},
  booktitle={International Conference on Learning Representations},
  volume={2025},
  pages={41488--41505},
  year={2025}
}

@article{chen2024magdi,
  title={Magdi: Structured distillation of multi-agent interaction graphs improves reasoning in smaller language models},
  author={Chen, Justin Chih-Yao and Saha, Swarnadeep and Stengel-Eskin, Elias and Bansal, Mohit},
  journal={arXiv preprint arXiv:2402.01620},
  year={2024}
}

@inproceedings{lee2025gemmas,
  title={Gemmas: Graph-based evaluation metrics for multi agent systems},
  author={Lee, Jisoo and Chang, Raeyoung and Kwon, Dongwook and Singh, Harmanpreet and Verma, Nikhil},
  booktitle={Proceedings of the 2025 Conference on Empirical Methods in Natural Language Processing: Industry Track},
  pages={1522--1532},
  year={2025}
}

@article{besta2025demystifying,
  title={Demystifying chains, trees, and graphs of thoughts},
  author={Besta, Maciej and Memedi, Florim and Zhang, Zhenyu and Gerstenberger, Robert and Piao, Guangyuan and Blach, Nils and Nyczyk, Piotr and Copik, Marcin and Kwa{\'s}niewski, Grzegorz and M{\"u}ller, Jurgen and others},
  journal={IEEE Transactions on Pattern Analysis and Machine Intelligence},
  year={2025},
  publisher={IEEE}
}

@article{wang2022self,
  title={Self-consistency improves chain of thought reasoning in language models},
  author={Wang, Xuezhi and Wei, Jason and Schuurmans, Dale and Le, Quoc and Chi, Ed and Narang, Sharan and Chowdhery, Aakanksha and Zhou, Denny},
  journal={arXiv preprint arXiv:2203.11171},
  year={2022}
}

@article{yao2023tree,
  title={Tree of thoughts: Deliberate problem solving with large language models},
  author={Yao, Shunyu and Yu, Dian and Zhao, Jeffrey and Shafran, Izhak and Griffiths, Tom and Cao, Yuan and Narasimhan, Karthik},
  journal={Advances in neural information processing systems},
  volume={36},
  pages={11809--11822},
  year={2023}
}

@inproceedings{qian2024chatdev,
  title={Chatdev: Communicative agents for software development},
  author={Qian, Chen and Liu, Wei and Liu, Hongzhang and Chen, Nuo and Dang, Yufan and Li, Jiahao and Yang, Cheng and Chen, Weize and Su, Yusheng and Cong, Xin and others},
  booktitle={Proceedings of the 62nd annual meeting of the association for computational linguistics (volume 1: Long papers)},
  pages={15174--15186},
  year={2024}
}

@article{dong2024self,
  title={Self-collaboration code generation via chatgpt},
  author={Dong, Yihong and Jiang, Xue and Jin, Zhi and Li, Ge},
  journal={ACM Transactions on Software Engineering and Methodology},
  volume={33},
  number={7},
  pages={1--38},
  year={2024},
  publisher={ACM New York, NY}
}

@article{talebirad2023multi,
  title={Multi-agent collaboration: Harnessing the power of intelligent llm agents},
  author={Talebirad, Yashar and Nadiri, Amirhossein},
  journal={arXiv preprint arXiv:2306.03314},
  year={2023}
}

@inproceedings{liang2024encouraging,
  title={Encouraging divergent thinking in large language models through multi-agent debate},
  author={Liang, Tian and He, Zhiwei and Jiao, Wenxiang and Wang, Xing and Wang, Yan and Wang, Rui and Yang, Yujiu and Shi, Shuming and Tu, Zhaopeng},
  booktitle={Proceedings of the 2024 conference on empirical methods in natural language processing},
  pages={17889--17904},
  year={2024}
}

@article{khan2024debating,
  title={Debating with more persuasive llms leads to more truthful answers},
  author={Khan, Akbir and Hughes, John and Valentine, Dan and Ruis, Laura and Sachan, Kshitij and Radhakrishnan, Ansh and Grefenstette, Edward and Bowman, Samuel R and Rockt{\"a}schel, Tim and Perez, Ethan},
  journal={arXiv preprint arXiv:2402.06782},
  year={2024}
}

@article{smit2023should,
  title={Should we be going mad? a look at multi-agent debate strategies for llms},
  author={Smit, Andries and Duckworth, Paul and Grinsztajn, Nathan and Barrett, Thomas D and Pretorius, Arnu},
  journal={arXiv preprint arXiv:2311.17371},
  year={2023}
}

@inproceedings{li2024improving,
  title={Improving multi-agent debate with sparse communication topology},
  author={Li, Yunxuan and Du, Yibing and Zhang, Jiageng and Hou, Le and Grabowski, Peter and Li, Yeqing and Ie, Eugene},
  booktitle={Findings of the Association for Computational Linguistics: EMNLP 2024},
  pages={7281--7294},
  year={2024}
}

@article{ishibashi2024self,
  title={Self-organized agents: A llm multi-agent framework toward ultra large-scale code generation and optimization},
  author={Ishibashi, Yoichi and Nishimura, Yoshimasa},
  journal={arXiv preprint arXiv:2404.02183},
  year={2024}
}

@article{zhang2024g,
  title={G-designer: Architecting multi-agent communication topologies via graph neural networks},
  author={Zhang, Guibin and Yue, Yanwei and Sun, Xiangguo and Wan, Guancheng and Yu, Miao and Fang, Junfeng and Wang, Kun and Chen, Tianlong and Cheng, Dawei},
  journal={arXiv preprint arXiv:2410.11782},
  year={2024}
}

@article{huang2026mmdeepresearch,
  title={MMDeepResearch-Bench: A Benchmark for Multimodal Deep Research Agents},
  author={Huang, Peizhou and Zhong, Zixuan and Wan, Zhongwei and Zhou, Donghao and Alam, Samiul and Wang, Xin and Li, Zexin and Dou, Zhihao and Zhu, Li and Xiong, Jing and others},
  journal={arXiv preprint arXiv:2601.12346},
  year={2026}
}

@article{wang2025fdabench,
  title={FDABench: A Benchmark for Data Agents on Analytical Queries over Heterogeneous Data},
  author={Wang, Ziting and Zhang, Shize and Yuan, Haitao and Zhu, Jinwei and Li, Shifu and Dong, Wei and Cong, Gao},
  journal={arXiv preprint arXiv:2509.02473},
  year={2025}
}

@inproceedings{liu2024agentbench,
  title={Agentbench: Evaluating llms as agents},
  author={Liu, Xiao and Yu, Hao and Zhang, Hanchen and Xu, Yifan and Lei, Xuanyu and Lai, Hanyu and Gu, Yu and Ding, Hangliang and Men, Kaiwen and Yang, Kejuan and others},
  booktitle={International Conference on Learning Representations},
  volume={2024},
  pages={52989--53046},
  year={2024}
}

@inproceedings{xu2024magic,
  title={Magic: Investigation of large language model powered multi-agent in cognition, adaptability, rationality and collaboration},
  author={Xu, Lin and Hu, Zhiyuan and Zhou, Daquan and Ren, Hongyu and Dong, Zhen and Keutzer, Kurt and Ng, See Kiong and Feng, Jiashi},
  booktitle={Proceedings of the 2024 Conference on Empirical Methods in Natural Language Processing},
  pages={7315--7332},
  year={2024}
}

@article{ma2024agentboard,
  title={Agentboard: An analytical evaluation board of multi-turn llm agents},
  author={Ma, Chang and Zhang, Junlei and Zhu, Zhihao and Yang, Cheng and Yang, Yujiu and Jin, Yaohui and Lan, Zhenzhong and Kong, Lingpeng and He, Junxian},
  journal={Advances in neural information processing systems},
  volume={37},
  pages={74325--74362},
  year={2024}
}

@inproceedings{wu2024smartplay,
  title={Smartplay: A benchmark for llms as intelligent agents},
  author={Wu, Yue and Tang, Xuan and Mitchell, Tom and Li, Yuanzhi},
  booktitle={International Conference on Learning Representations},
  volume={2024},
  pages={1543--1561},
  year={2024}
}

@article{wang2026trajectory2task,
  title={Trajectory2Task: Training Robust Tool-Calling Agents with Synthesized Yet Verifiable Data for Complex User Intents},
  author={Wang, Ziyi and Lu, Yuxuan and Zhang, Yimeng and Chen, Pei and Dong, Ziwei and Huang, Jing and Gesi, Jiri and Tang, Xianfeng and Luo, Chen and Liu, Qun and others},
  journal={arXiv preprint arXiv:2601.20144},
  year={2026}
}

@article{wang2026aligning,
  title={Aligning Agents via Planning: A Benchmark for Trajectory-Level Reward Modeling},
  author={Wang, Jiaxuan and Hu, Yulan and Yang, Wenjin and Pan, Zheng and Li, Xin and Guo, Lan-Zhe},
  journal={arXiv preprint arXiv:2604.08178},
  year={2026}
}

@article{zhang2024chain,
  title={Chain of agents: Large language models collaborating on long-context tasks},
  author={Zhang, Yusen and Sun, Ruoxi and Chen, Yanfei and Pfister, Tomas and Zhang, Rui and Ar{\i}k, Sercan {\"O}},
  journal={Advances in Neural Information Processing Systems},
  volume={37},
  pages={132208--132237},
  year={2024}
}

@article{pan2025agentcoord,
  title={Agentcoord: Visually exploring coordination strategy for llm-based multi-agent collaboration},
  author={Pan, Bo and Lu, Jiaying and Wang, Ke and Zheng, Li and Wen, Zhen and Feng, Yingchaojie and Zhu, Minfeng and Chen, Wei},
  journal={Computers \& graphics},
  pages={104338},
  year={2025},
  publisher={Elsevier}
}

@inproceedings{wang2025agentdropout,
  title={Agentdropout: Dynamic agent elimination for token-efficient and high-performance llm-based multi-agent collaboration},
  author={Wang, Zhexuan and Wang, Yutong and Liu, Xuebo and Ding, Liang and Zhang, Miao and Liu, Jie and Zhang, Min},
  booktitle={Proceedings of the 63rd Annual Meeting of the Association for Computational Linguistics (Volume 1: Long Papers)},
  pages={24013--24035},
  year={2025}
}

@article{li2025adaptive,
  title={Adaptive graph pruning for multi-agent communication},
  author={Li, Boyi and Zhao, Zhonghan and Lee, Der-Horng and Wang, Gaoang},
  journal={arXiv preprint arXiv:2506.02951},
  year={2025}
}

@inproceedings{shen2025understanding,
  title={Understanding the information propagation effects of communication topologies in llm-based multi-agent systems},
  author={Shen, Xu and Liu, Yixin and Dai, Yiwei and Wang, Yili and Miao, Rui and Tan, Yue and Pan, Shirui and Wang, Xin},
  booktitle={Proceedings of the 2025 Conference on Empirical Methods in Natural Language Processing},
  pages={12358--12372},
  year={2025}
}

@article{michelakis2025core,
  title={CORE: Full-Path Evaluation of LLM Agents Beyond Final State},
  author={Michelakis, Panagiotis and Hadjiyiannis, Yiannis and Stamoulis, Dimitrios},
  journal={arXiv preprint arXiv:2509.20998},
  year={2025}
}

@inproceedings{ou2025agentdiagnose,
  title={AgentDiagnose: An Open Toolkit for Diagnosing LLM Agent Trajectories},
  author={Ou, Tianyue and Guo, Wanyao and Gandhi, Apurva and Neubig, Graham and Yue, Xiang},
  booktitle={Proceedings of the 2025 Conference on Empirical Methods in Natural Language Processing: System Demonstrations},
  pages={207--215},
  year={2025}
}

@article{du2023improving,
  title={Improving factuality and reasoning in language models through multiagent debate},
  author={Du, Yilun and Li, Shuang and Torralba, Antonio and Tenenbaum, Joshua B and Mordatch, Igor},
  journal={arXiv preprint arXiv:2305.14325},
  year={2023}
}

@article{ye2025mas,
  title={Mas-gpt: Training llms to build llm-based multi-agent systems},
  author={Ye, Rui and Tang, Shuo and Ge, Rui and Du, Yaxin and Yin, Zhenfei and Chen, Siheng and Shao, Jing},
  journal={arXiv preprint arXiv:2503.03686},
  year={2025}
}

@misc{openai2024swarm,
  title        = {Swarm: An Educational Framework Exploring Ergonomic, Lightweight Multi-Agent Orchestration},
  author       = {{OpenAI}},
  year         = {2024},
  howpublished = {\url{https://github.com/openai/swarm}},
  note         = {GitHub repository}
}

@misc{openai2026introducing,
  author       = {{OpenAI}},
  title        = {Introducing GPT-5.5},
  howpublished = {\url{https://openai.com/index/introducing-gpt-5-5/}},
  year         = {2026},
  month        = {April},
}

@misc{deepseekai2026deepseekv4,
      title={DeepSeek-V4: Towards Highly Efficient Million-Token Context Intelligence},
      author={DeepSeek-AI},
      year={2026},
}

@misc{anthropic2026sonnet46,
  author       = {{Anthropic}},
  title        = {Introducing Sonnet 4.6},
  howpublished = {\url{https://www.anthropic.com/news/claude-sonnet-4-6}},
  year         = {2026},
  month        = {February},
}

@misc{qwen36plus,
    title = {{Qwen3.6-Plus}: Towards Real World Agents},
    url = {https://qwen.ai/blog?id=qwen3.6},
    author = {{Qwen Team}},
    month = {April},
    year = {2026}
}

\clearpage
\appendix
\section{Supplementary Metric Validation}
\label{app:supplementary-validation}

We report the full supporting experiments for the metric-behavior summary in Section~\ref{sec:metric-behavior}. Five controlled perturbations are applied at $p\in\{0.3,0.5\}$ to a stratified sample. Four structural perturbations---node deletion, edge rewiring, role swapping, and distractor insertion---reduce Forest Match as intended. Duplicate-content injection leaves the set-based Forest Match nearly unchanged and is instead captured by Content Redundancy.

\begin{table}[h]
\centering
\small
\caption{Structure-level perturbation validation on a stratified sample.}
\label{tab:structure-perturb}
\resizebox{\linewidth}{!}{
\begin{tabular}{lcc}
\toprule
Reported quantity & Value & Setting \\
\midrule
Settings with Kendall $\tau=1.000$ & 9/10 & five perturbations $\times$ two intensities \\
Minimum Kendall $\tau$ & 0.867 & worst perturbation setting \\
Largest FM decrease & $-0.058$ & node deletion at $p=0.5$ \\
Redundancy-injection $\Delta\mathrm{FM}$ & $\approx0$ & $p\in\{0.3,0.5\}$ \\
\bottomrule
\end{tabular}
}
\end{table}

We additionally re-aggregate Forest Match under multiple weight schemes and sweep 36 combinations of $\theta_d\in\{4,5,6\}$, $\theta_w\in\{0.4,0.5,0.6\}$, and $\theta_\delta\in\{-0.5,0.0,0.5,1.0\}$. Table~\ref{tab:param-sensitivity} shows that the main ordering is stable across non-degenerate weight schemes and all tested TSP thresholds. The topology-only case is intentionally outside the metric's intended use because it removes role and edge semantics.

\begin{table}[h]
\centering
\small
\caption{Sensitivity of framework ranking to Forest Match weights and TSP thresholds. The threshold grid retains $222$--$1{,}244$ tasks depending on the setting.}
\label{tab:param-sensitivity}
\resizebox{\linewidth}{!}{
\begin{tabular}{lccc}
\toprule
Sensitivity axis & Settings & Kendall $\tau$ & Rank preservation \\
\midrule
Non-degenerate FM weights & 5 & $\ge0.73$ & Top-3: 5/5 \\
Uniform/node/edge-heavy & 3 & $\ge0.87$ & Top-3: 3/3 \\
Topology-only weight & 1 & $0.20$ & outside intended use \\
TSP threshold grid & 36 & $1.00$ in 36/36 & Top-1: 36/36 \\
\bottomrule
\end{tabular}
}
\end{table}

\section{Candidate-Backbone Analysis}
\label{app:backbone-analysis}

We first represent each trace by graph topology, role mixture, motif identity, and reference proximity, and compare traces produced by the same framework under different candidate backbones. The aggregate cross-backbone distance is $0.171$, which is slightly below the within-backbone distance of $0.181$. The three framework-by-framework distance matrices correlate at Pearson $0.986$--$0.999$. This supports the conclusion that the framework imposes a strong structural prior across the tested backbones.

\begin{figure*}[t]
\centering
\begin{tcolorbox}[
  width=0.96\textwidth,
  colback=blue!5!gray!5,
  colframe=blue!30!gray,
  colbacktitle=blue!15!gray!15,
  boxrule=0.3mm, arc=2mm,
  left=6pt, right=6pt, top=5pt, bottom=5pt,
  title={Failure-diagnostic judge prompt variants},
  fonttitle=\bfseries\footnotesize\color{black},
  coltitle=black
]
\footnotesize
\textbf{(a) Original rubric} \hfill \textbf{$\sim$1,466 characters}
\vspace{1pt}\hrule\vspace{3pt}
{\itshape You are a multi-agent-system (MAS) failure analyst. Given a TASK, a GOLD answer, a reference trace from the reference forest that solved it correctly, and a CANDIDATE run from a specific framework, classify how the candidate failed (or succeeded). Output STRICT JSON only with keys \texttt{structural\_failure}, \texttt{content\_failure}, \texttt{error\_first\_at}, \texttt{was\_caught}, \texttt{final\_writer}, \texttt{brief\_explanation}, and \texttt{min\_fix\_hint}. The rubric defines structural labels such as \texttt{missing\_role}, \texttt{wrong\_motif}, \texttt{truncated\_pipeline}, \texttt{redundant\_loop}, \texttt{no\_aggregation}, and \texttt{ok}, and content labels such as \texttt{wrong\_decomposition}, \texttt{hallucinated\_fact}, \texttt{aggregator\_picks\_wrong}, \texttt{format\_error}, and \texttt{ok}. It also includes short rules clarifying when \texttt{wrong\_motif} and \texttt{missing\_role} apply.}

\vspace{6pt}
\textbf{(b) Concise rubric} \hfill \textbf{$\sim$490 characters}
\vspace{1pt}\hrule\vspace{3pt}
{\itshape You are a MAS failure analyst. Output STRICT JSON only with the same schema and label set as above. No prose outside JSON. This variant keeps the output schema but removes the per-key explanations, examples, and specificity nudges, testing whether the taxonomy is stable when the rubric wording is compressed.}

\vspace{6pt}
\textbf{(c) Few-shot rubric} \hfill \textbf{$\sim$1,850 characters}
\vspace{1pt}\hrule\vspace{3pt}
{\itshape This variant keeps the original schema and adds two worked examples: one failed debate-style trace labelled with \texttt{no\_aggregation}/\texttt{aggregator\_picks\_wrong}, and one successful planner-solver-verifier trace labelled \texttt{ok}/\texttt{ok}. It tests whether adding examples stabilizes the diagnostic labels without changing the task inputs.}

\vspace{6pt}
\textbf{Shared user template.} All three prompt variants receive the same per-trace user message containing task metadata, accuracy, Forest Match, token cost, the query, the gold answer, the candidate final answer, a reference-trace block from the reference forest, and the candidate run block. Trace step content is capped identically across variants.
\end{tcolorbox}
\caption{System-prompt variants for the prompt-variant robustness check. The full experiment runs the same DeepSeek judge on the same $100$ traces with these three rubric variants; Table~\ref{tab:prompt-variant-judge} reports the resulting pairwise agreement.}
\label{box:judge-prompts}
\end{figure*}
We then compare each framework--backbone pair with an unguided single-agent chain-of-thought baseline on the same 60-query intersection. Table~\ref{tab:deltaacc} shows that the marginal accuracy gain of orchestration depends on the backbone and the framework; the average gain becomes negative for the stronger tested backbones even though individual pairs can improve.

\begin{table}[h]
\centering
\small
\caption{Accuracy gain $\Delta\mathrm{acc}$ over the single-agent CoT baseline on the same backbone and 60-query intersection.}
\label{tab:deltaacc}
\begin{tabular}{lccc}
\toprule
Backbone & Mean $\Delta$ & Best $\Delta$ & Worst $\Delta$ \\
\midrule
DS-V4 & $-0.002$ & $+0.042$, MAS-GPT & $-0.050$, AFlow \\
GPT-5.5 & $-0.039$ & $+0.070$, Debate & $-0.129$, AFlow \\
Claude-S4.6 & $-0.069$ & $\phantom{-}0.000$, Swarm & $-0.150$, AFlow \\
\bottomrule
\end{tabular}
\end{table}

\section{Failure Diagnostic Details}
\label{app:failure-diagnostics}

In this section, we provide the supporting protocol details for the failure diagnostics. We sample $300$ candidate traces stratified by framework, dataset, and the accuracy--Forest Match quadrant. Each diagnostic input contains the task, gold answer, candidate answer, one successful reference run, the candidate trace, accuracy, Forest Match, and token cost. The diagnostic outputs structural and content labels, the first error location, whether a later agent caught the error, and a minimal repair suggestion.

We run DeepSeek-V4-Flash, GPT-5.5, and Qwen3.6-plus and aggregate labels by majority vote. Because GPT-5.5 produces parse-clean JSON on only a subset, Table~\ref{tab:judge-panel} reports the shared parse-clean intersection with $n=138$. The analysis is used only to support coarse structural-versus-content repair directions; it is not part of the benchmark score.

\section{Diagnostic Robustness}
\label{app:diagnostic-scope}

We additionally run the same DeepSeek-V4-Flash judge on the same $100$ traces with three rubric variants. Table~\ref{tab:prompt-variant-judge} shows that fine-grained labels are sensitive to rubric wording, particularly for structural failures. This motivates the main text's emphasis on coarse repair directions rather than exact label-frequency claims.

\begin{table}[h]
\centering
\small
\caption{Prompt sensitivity of the LLM failure diagnostic. Values are pairwise Cohen's $\kappa$ ranges across three rubric variants on the same $100$ traces.}
\label{tab:prompt-variant-judge}
\begin{tabular}{lc}
\toprule
Diagnostic axis & Cohen's $\kappa$ range \\
\midrule
Structural failure & 0.129--0.351 \\
Content failure & 0.331--0.470 \\
Whether the error was caught & 0.388--0.621 \\
First error location & 0.537--0.616 \\
\bottomrule
\end{tabular}
\end{table}

\section{Additional Reference-Forest Robustness}
\label{app:reference-forest-robustness}

Table~\ref{tab:leave-source-out} reports the source-level reference-forest robustness experiment. The main text reports leave-framework-out ablations; this appendix adds the orthogonal leave-source-out view. The leave-one-source-out setting removes one dataset source at a time and ranks frameworks on the remaining pool, while the per-source setting ranks frameworks on each single source separately.

\begin{table}[h]
\centering
\small
\caption{Source-level reference-forest robustness. The leave-one-source-out setting tests whether the aggregate ranking depends on any single dataset source; the per-source setting exposes expected domain-specific specialization.}
\label{tab:leave-source-out}
\resizebox{\linewidth}{!}{
\begin{tabular}{lccc}
\toprule
Setting & Sources & Kendall $\tau$ vs. full ranking & Rank preservation \\
\midrule
Leave-one-source-out & 8 & $0.87$--$1.00$ & Top-1: 8/8 \\
Per-source ranking & 7 with $\ge20$ traces & $-0.33$--$1.00$ & Top-1: 4/7 \\
SWE-bench-V only & 1 & $-0.33$ & Debate ranks first \\
\bottomrule
\end{tabular}
}
\end{table}

The leave-one-source-out result which includes CoT-SC shows that no single dataset source dominates the aggregate ranking. The per-source result is intentionally more variable: individual domains can favor different collaboration motifs. In particular, the SWE-bench result is consistent with the main-text observation that Debate's review-and-revise motif is unusually well matched to code-patch style tasks.

\end{document}